\documentclass[11pt]{article}

\usepackage[margin=1in]{geometry}

\usepackage[T1]{fontenc}

\usepackage[utf8]{inputenc}

\usepackage{lmodern}

\usepackage{setspace}

\usepackage{enumitem}

\usepackage{hyperref}

\usepackage{titlesec}

\usepackage{graphicx}
\usepackage[most]{tcolorbox}

\newtcolorbox{journalbox}[1]{
colback=gray!10,
colframe=gray!45,
boxrule=0.4pt,
arc=2pt,
left=8pt,
right=8pt,
top=8pt,
bottom=8pt,
title={#1},
fonttitle=\bfseries,
coltitle=black,
colbacktitle=gray!18,
attach boxed title to top left={xshift=0.5em,yshift=-2mm},
boxed title style={boxrule=0pt,arc=1pt,colframe=gray!18,colback=gray!18},
before skip=10pt,
after skip=10pt
}

\setlist[itemize]{leftmargin=1.5em}

\setlist[enumerate]{leftmargin=1.8em}

\titleformat{\section}{\large\bfseries}{\thesection.}{0.5em}{}

\titleformat{\subsection}{\normalsize\bfseries}{\thesubsection.}{0.5em}{}

\title{\textbf{From Prediction to Action: Population Health World Model for Digital Medicine}}

\author{%
\parbox{\textwidth}{\raggedright
Xike Xie$^{1-3}$, Yu Li$^{6-8}$, Weizhong Yang$^{6-8}$, Amalie Dyda$^9$, and Shaohua Kevin Zhou$^{1-5}$\thanks{Corresponding author: skevinzhou@ustc.edu.cn} \\[8pt]
{\small \it
$^1$School of Biomedical Engineering, Division of Life Sciences and Medicine, University of Science and Technology of China, Hefei, Anhui 230026, China\\
$^2$Yangtze River Delta Research Institute of Group Health Technology - MIRACLE Lab, USTC Suzhou Institute for Advanced Research, Suzhou, Jiangsu 215123, China\\
$^3$Jiangsu Provincial Key Laboratory of Multimodal Digital Twin Technology, Suzhou, Jiangsu 215123, China\\
$^4$Biomedical Basic Research Center of Jiangsu Province, Suzhou, Jiangsu 215123, China\\
$^5$State Key Laboratory of Precision and Intelligent Chemistry, Hefei, Anhui 230026, China\\
$^6$School of Population Medicine and Public Health, Chinese Academy of Medical Sciences \& Peking Union Medical College, Beijing 100006, China\\
$^7$Innovation Center for Philosophy and Social Sciences in Higher Education, Innovation Center for Public Health Emergency Management, Beijing 100006, China\\
$^8$State Key Laboratory of Respiratory Health and Multimorbidity, Beijing 100006, China\\
$^9$School of Public Health, The University of Queensland, Brisbane, Queensland, Australia} 
}
}% author block ending “}”

\begin{document}
\maketitle

\begin{center}
\small\textit{An earlier version of this manuscript was first
submitted for journal publication on May 15, 2026.}
\end{center}

\begin{abstract} Digital medicine now captures data that traces individual health trajectories and reveals population health dynamics through electronic health records, claims, remote monitoring, wearable devices, mobile devices, social sensing data, medical images, etc. But much of its modeling still centers on task-specific, outcome-specific, and prediction-specific estimation. The next frontier is not simply better prediction, but action in an evolving health world shaped by multiple decision-makers, heterogeneous populations, resource constraints, and feedback in care delivery. This is why digital medicine now needs world models: not only to estimate what will happen, but to represent how actions change what can happen next -- whom to prioritize, how to allocate scarce resources, how to adapt care across subgroups, and how interventions reshape future health trajectories and population dynamics.

We call such a model \textbf{population health world model (PHWM)}: a more foundational representation of an evolving health world in which health states, individual trajectories, population health dynamics, interventions, constraints, and feedback are intrinsically coupled. Building a PHWM endows three unmet structural gaps in current digital medicine: dynamic health-state modeling, population-level intervention intelligence, and validation, equity, and governance for action-oriented use. 
We argue that addressing these gaps requires three minimal capabilities: representation learning under action-dependent observation, multi-scale intervention intelligence, and validation with explicit decision boundaries, while highlighting the importance of social, contextual, and public-health sensing data for action-oriented digital medicine.
\end{abstract}

%\section*{Main}

\subsection*{Prediction Is No Longer Enough}

Digital medicine now observes far more than episodic clinical events. Electronic health records, claims, remote monitoring, wearable sensors, medical imaging, etc. %mobile applications, and patient-reported data 
increasingly capture how health evolves within individuals and across populations \cite{doi:10.1126/scitranslmed.abn6036,Low2020CancerSensors,Ghomrawi2023WearablePostOp,Mahajan2025WearableAI}. In practice, these richer observations have mostly been used as an expanded input space for task-specific prediction, including readmission, hospitalization, treatment escalation, or which alert should fire next \cite{Moor2023GeneralistMedicalAI,Kalmady2024PopulationECG}. That framing fits many established digital medicine tasks. But the frontier is shifting from prediction to action: whom to prioritize, how to allocate scarce resources, how to adapt care across subgroups, and how interventions reshape future health trajectories and population health dynamics.

\begin{figure}

\centering

\includegraphics[width=0.95\linewidth]{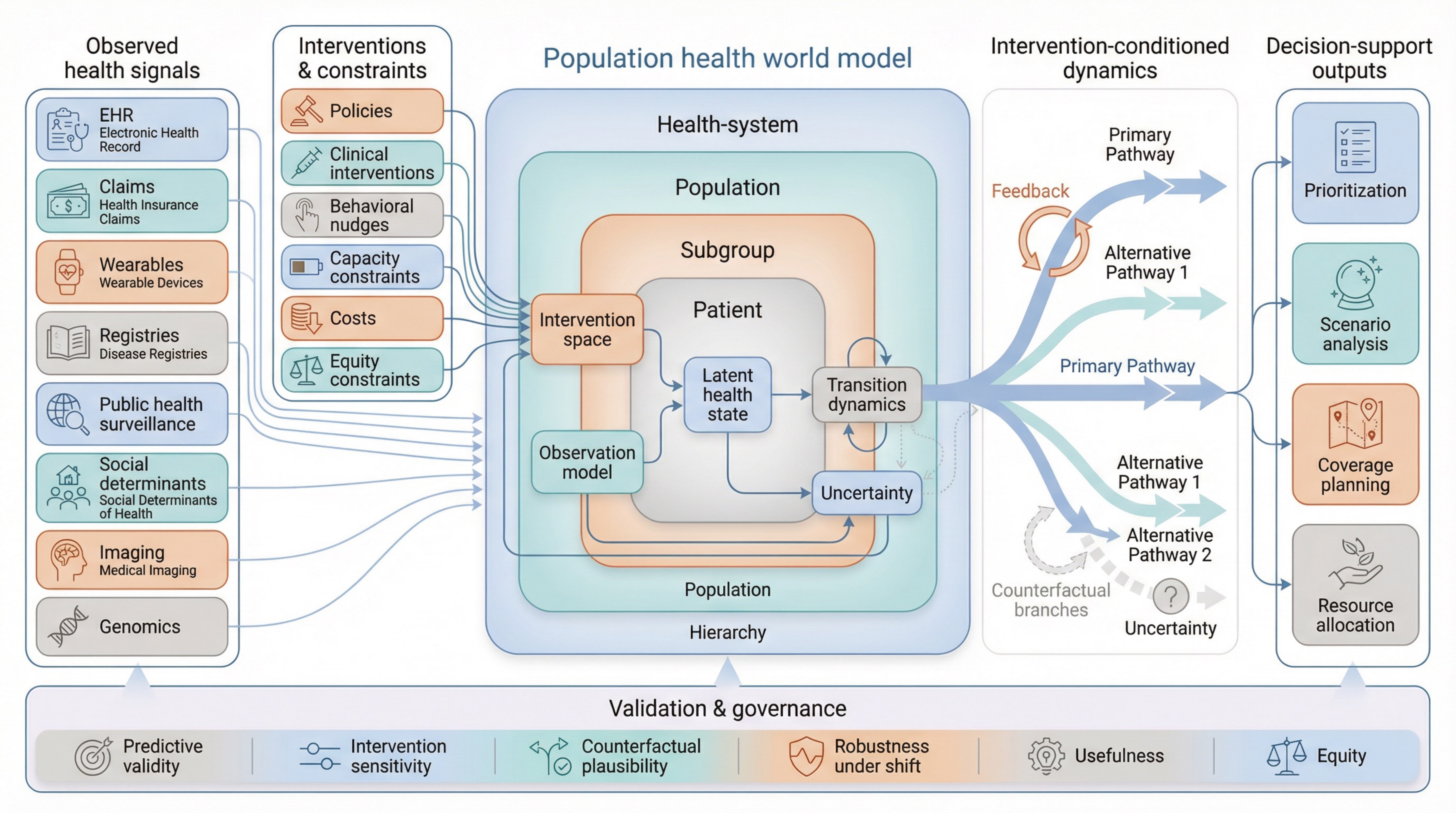}

\caption{{\bf Overview of Population Health World Model:} \small Observed health signals---including electronic health records, claims, wearable and remote-monitoring data, patient-reported outcomes, and social or contextual signals---feed into a population health world model that distinguishes latent health state from recorded observations. The model represents intervention space, transition dynamics, uncertainty, and hierarchy across levels and time scales, linking patient-level health trajectories to subgroup patterns and population health dynamics. Interventions and real-world constraints jointly shape intervention-conditioned dynamics, which in turn support action-oriented outputs such as prioritization, scenario analysis, coverage planning, and resource allocation. Validation and governance form an integrated layer spanning the framework, emphasizing predictive validity, intervention sensitivity, counterfactual plausibility, robustness under shift, usefulness, and equity.}

\label{fig:phwm}
\end{figure}

Action changes the health world being modeled. Consequently, new action workflows change who is measured, when follow-up occurs, how adherence evolves, and where capacity is consumed. Once action changes the care process, prediction is no longer enough. A model that scores risk under yesterday's workflow does not tell us whom to prioritize today, how to act under limited capacity, or who will be missed or burdened when a policy is put into practice. Therefore, digital medicine needs more than better prediction. It needs a model that represents how interventions reshape health trajectories, observations, and population dynamics. Figure~\ref{fig:phwm} summarizes this modeling perspective and presents an overview of a Population Health World Model (PHWM).

\subsection*{What a Population Health World Model Is}

A PHWM is not simply a stronger predictor with more inputs, nor a standard prediction model with an intervention variable added on top. It is a model of an evolving health world in which individual health trajectories, population health dynamics, interventions, constraints, and feedback are intrinsically coupled \cite{LeCun2022PathAMI,Ha2018WorldModels}. 

A broader health world extends beyond formal healthcare. Digital and public-health sensing data---including search behavior, social media, mobility, wastewater surveillance, environmental exposure, and community-level socioeconomic indicators---can reveal population-level demand, risk perception, behavior, exposure, and access constraints that are often invisible in clinical records alone.

%A population health world model is not simply a stronger predictor with more inputs, nor a standard prediction model with an intervention variable added on top. 
%Beyond formal healthcare, digital and public-health sensing data---including search behavior, social media, mobility, wastewater surveillance, environmental exposure, and community-level socioeconomic indicators---can reveal population-level demand, risk perception, behavior, exposure, and access constraints.
%It is a model of an evolving health world: one in which individual health trajectories, population health dynamics, interventions, constraints, and feedback are intrinsically coupled \cite{LeCun2022PathAMI,Ha2018WorldModels}.

That distinction matters because action does not just change outcomes. It changes observation, follow-up, adherence, capacity, and the downstream conditions under which future decisions are made. A model for action must therefore represent not only what is likely to happen, but how actions change what can happen next.

\begin{journalbox}{Box 1 | Working Definition}
A \textbf{population health world model (PHWM)} is a computational model that represents evolving health states across individuals and populations together with interventions, context, and feedback over time to support prediction, counterfactual reasoning, and decision-making in digital medicine.
\end{journalbox}

At minimum, such a model should represent six core elements \cite{LeCun2022PathAMI,Ha2018WorldModels,Kaelbling1998POMDP,Mao2024GraphFoundationModels,DBLP:conf/nips/HaS18}.

First, \textbf{latent health state} rather than labels, capturing underlying disease, treatment, behavior, and context across individuals, subgroups, and care settings.

Second, an explicit \textbf{observation model}, since what is measured depends on care processes, recorded interactions, and event timing and cannot be equated with state.

Third, \textbf{interventions or actions} as part of the model itself, not an add-on to prediction, because actions modify not only patient status but also patterns of monitoring, contact, and care delivery.

Fourth, \textbf{transition dynamics} that describe how health states evolve under intervention, context, and constraints and how effects propagate across individuals, subgroups, and care settings.

Fifth, \textbf{uncertainty over partially observed state and future evolution}, reflecting limited observation and multiple plausible futures.

Sixth, \textbf{hierarchy across levels and time scales}, linking individual trajectories, subgroup patterns, population dynamics, and the care processes that connect them.

On top of these elements, digital medicine also requires explicit \textbf{objectives and constraints}, including effectiveness, burden, safety, efficiency, and equity. These objectives operate over individuals, subgroups, and populations simultaneously and are shaped by how care is delivered under real-world constraints.

\begin{figure}[t]
\centering
\includegraphics[width=0.9\linewidth]{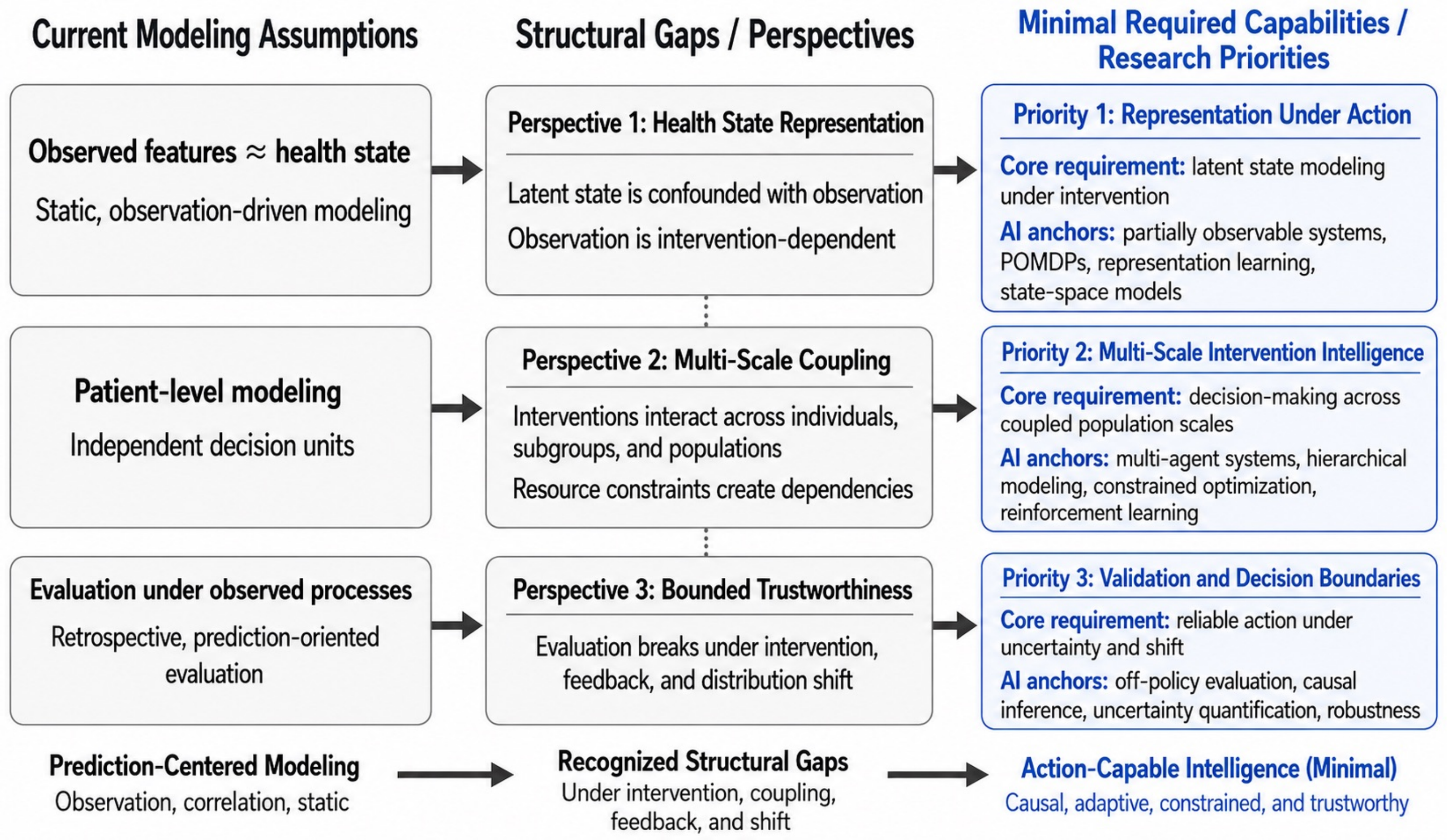}
\caption{\textbf{Structural Gaps and Minimal Required Capabilities in Population Health World Model.}
\small Current digital medicine models rely on assumptions that observed features approximate health state, that decisions can be modeled at the patient level, and that evaluation under observed processes is sufficient. These assumptions lead to three structural gaps: (1) representational challenges arising from latent state being confounded with intervention-dependent observations, (2) multi-scale coupling challenges due to interactions across individuals, subgroups, and populations under shared constraints, and (3) validation challenges under intervention, feedback, and distribution shift. Addressing these gaps requires minimal capabilities rather than optional directions: representation learning under action-dependent observation, multi-scale intervention intelligence, and validation frameworks with explicit decision boundaries. These priorities correspond to core problem classes in artificial intelligence, including partially observable modeling, multi-agent and hierarchical systems, and off-policy and uncertainty-aware evaluation.}
\label{fig:agenda}
\end{figure}

This framing also clarifies what a PHWM is not. It is not a standard predictive model that estimates outcomes under observed processes; not a digital twin focused mainly on individual fidelity; and not a causal estimator or sequential decision system alone. \textit{It is a model of an evolving and interconnected health world that links patient-level trajectories, population-level consequences, and action under constraints.}
Figure~\ref{fig:agenda} summarizes how these challenges arise from limitations in the current predictive paradigm, forming new perspectives, and why the priorities below should be understood as minimal required capabilities rather than optional research directions.
% =========================
% Figure 2 (agenda / diagnostic)
% =========================

\subsection*{Perspective 1: From Static Predictors to Dynamic Health-State Models}

The first perspective is that health state must be modeled, not merely predicted. The transition from static prediction to dynamic health-state models requires learning a representation of the clinically relevant state that is not directly observed \cite{Kaelbling1998POMDP,LeCun2022PathAMI}.
Static predictors map recorded inputs to predefined outcomes. Dynamic health-state models must do something harder: infer latent health state from measurements, encounters, claims, and interactions produced through care processes, and keep that representation meaningful when intervention changes both what is observed and what can happen next.

Consider blood-pressure management. The same observed pattern---for example, a missing home reading followed by no medication change---can correspond to very different underlying states: clinical stability, poor adherence, device failure, loss to follow-up, or lack of outreach. Those possibilities matter because they imply different next actions. They also imply different futures: intensifying monitoring, adjusting medication, or doing nothing will change not only the patient's subsequent health trajectory, but also what data are observed next. The representational problem is therefore to infer the latent health state that explains incomplete observations while preserving the action-dependent structure of future health trajectories and observations.

This is technically difficult because health observations are sparse, irregular, heterogeneous, and shaped by prior action(s). Electronic health records, wearable data, mobile applications, claims, and contextual signals each provide incomplete and differently biased views of the same evolving health world \cite{Low2020CancerSensors,Ghomrawi2023WearablePostOp,Gupta2023TrajectoryPatterns,Bladon2025PassiveData,Lampos2021OnlineSearch}. Thus, a PHWM must connect these observations to latent health state without a collapse due to the partial observability, structured dependencies, and cross-channel heterogeneity that matter for intervention.

The field therefore needs a stronger commitment to modeling health trajectories under action, not only labels under observation. Short-horizon predictive performance remains valuable, but it is not sufficient evidence that a model captures intervention-aware health dynamics.

\subsection*{Perspective 2: From Observed Health Signals to Multi-Scale Intervention Intelligence}

The second perspective is that population health intelligence is not scale alone, but multi-scale coupling. The move from dynamic health-state models to population intervention intelligence requires models that do not stop at individual health trajectories, but connect them to subgroup patterns, population health dynamics, and care-delivery constraints \cite{Richardson2022DigitalHealthEquity,Wilson2024AdvanceDigitalHealthEquity,Wang2025ChinaGovernance}. This is difficult because the consequence of an action is rarely confined to one patient or one group: interventions redistribute attention, consume capacity, and reshape what can be done elsewhere in the population.

Consider a vaccination or screening campaign. Not all groups can be reached at once, so programs must decide where to begin, whom to prioritize, and how to expand coverage over time. These decisions do not only affect the groups reached first. They determine which segments of the population are observed, protected, or left at risk, and their effects unfold as coverage expands. The technical problem is therefore not simply to rank individuals by need, but to represent how interventions distributed across groups reshape population health dynamics under shared constraints.

This is difficult because the relevant signals and consequences live at different levels. Individual-level data describe symptoms, adherence, and utilization. Subgroup structure captures systematic differences in engagement, access, and response. Population-level consequences include coverage, unmet need, downstream demand, and disparities in who benefits from intervention \cite{Richardson2022DigitalHealthEquity,Wilson2024AdvanceDigitalHealthEquity,Ji2025EquityGuard}. A useful PHWM must therefore couple individual health trajectories with subgroup structure, population health dynamics, and operational constraints rather than treat the population as a simple aggregate of isolated patients.

A PHWM should therefore support multi-scale intervention intelligence. It should represent how actions propagate across individuals, subgroups, and populations under real constraints. This does not require modeling everything. It requires modeling enough of the health world that decisions remain meaningful beyond one health trajectory at a time.

\subsection*{Perspective 3: Validation, Equity, and Governance}

The third perspective is that action-oriented models require bounded trustworthiness, not retrospective accuracy alone.
Once a model is used to support intervention, evaluation can no longer stop at predictive accuracy on historical data \cite{doi:10.1126/scitranslmed.abn6036,Ranisch2025SocialExperiment,Wang2025ChinaGovernance}. The key question becomes whether the model remains trustworthy when its outputs influence who receives attention, how care is adapted, and how downstream health trajectories unfold.

Consider a program that prioritizes outreach for vaccination, screening, or chronic-disease follow-up. It may perform well on retrospective data yet still fail in practice: the outreach it triggers can change subsequent observations, alter uptake and care utilization, and systematically underserve groups that are harder to measure or engage. The problem is therefore not only whether the model predicts well, but whether it remains credible, useful, and fair once deployed for action.

This is why validation must be broader than conventional predictive benchmarking. A PHWM should be evaluated along at least six dimensions: predictive validity, intervention sensitivity, counterfactual plausibility, robustness under shift, clinical or operational usefulness, and equity \cite{Ranisch2025SocialExperiment,Wilson2024AdvanceDigitalHealthEquity,Ji2025EquityGuard}. Governance follows naturally from the same concern. These models may integrate sensitive data, include opaque components, and shape how outreach, coverage, or intensified care are distributed across populations. Their role must therefore be bounded, their intended uses explicit, and their oversight built into the workflow rather than added after deployment.

The goal is not to certify a world model as universally correct. It is to establish bounded trustworthiness for specific action settings. Human oversight is therefore not an afterthought, but part of the socio-technical environment in which the model operates.

Together, these perspectives identify structural gaps in the current predictive paradigm. They do not imply a single implementation pathway. Instead, they motivate three minimal capabilities required for population health world models to support action-oriented digital medicine.

\subsection*{Priority 1: Representation Learning Under Action}

The first priority is to establish a minimal representation capable of supporting reasoning under intervention. Technically, this means learning a latent health state from partially observed, irregular, and intervention-shaped data streams while exposing the interfaces needed for observation models, intervention space, transition dynamics, uncertainty over hidden state and future evolution, and hierarchy across levels and time scales \cite{LeCun2022PathAMI,Ha2018WorldModels,Kaelbling1998POMDP,Mao2024GraphFoundationModels}.
%The first priority is to define the core abstractions of the health world under action. Technically, this means specifying the minimal representational objects and interfaces that a population health world model must expose: latent health state, observation model, intervention space, transition dynamics, uncertainty over hidden state and future evolution, and hierarchy across levels and time scales \cite{LeCun2022PathAMI,Ha2018WorldModels,Kaelbling1998POMDP,Mao2024GraphFoundationModels}.

A useful abstraction must enforce explicit separation between latent health state and observed data, and must model dynamics as conditioned on interventions. Concretely, this requires: (i) a state representation that is not reducible to recorded features alone, (ii) an observation model that accounts for missingness, irregular sampling, and measurement bias, and (iii) transition operators that encode how interventions alter both future state and future observations.

The abstraction must also specify identifiability and compositional structure. Under partial and action-dependent observation, the model should make explicit which aspects of state and dynamics are learnable, what invariances are assumed across subgroups and populations, and how individual health trajectories compose into subgroup patterns and population health dynamics. Objectives and constraints should be defined separately as part of the decision layer built on top of the world model, rather than conflated with the representation itself.

\subsection*{Priority 2: Multi-Scale Intervention Intelligence}

The second priority is to enable decision-making across coupled population scales under resource and system constraints. Technically, this requires task formulations and benchmarks that evaluate whether a model supports prioritization, coverage allocation, dynamic resource allocation, and subgroup-aware intervention when actions propagate across individuals, subgroups, and populations \cite{Valmeekam2023PlanBench,Qiao2024WorldKnowledgeModel}.
%The second priority is to build benchmarks for multi-scale intervention intelligence. Technically, such benchmarks should evaluate whether a model supports prioritization, coverage allocation, dynamic resource allocation, and subgroup-aware intervention under explicit constraints \cite{Valmeekam2023PlanBench,Qiao2024WorldKnowledgeModel}.

A useful benchmark framework should specify: (i) the intervention space being evaluated, such as outreach, coverage expansion, monitoring intensity, or resource allocation; (ii) the levels at which consequences are measured, including individual health trajectories, subgroup patterns, and population health dynamics; and (iii) the constraints under which decisions are made, such as limited capacity, staged rollout, or unequal observability across groups.

Benchmark tasks should therefore make multi-scale coupling visible rather than average it away. Useful settings include intervention scenario forecasting, staged coverage expansion, constrained outreach allocation, subgroup-sensitive stress tests, and evaluation of how benefit, burden, and unmet need are redistributed across populations. The goal is to measure intervention intelligence rather than patient ranking alone.

\subsection*{Priority 3: Validation and Decision Boundaries}

The third priority is to formalize when model outputs are reliable enough to support action under intervention, uncertainty, and distribution shift. Technically, this requires specifying the mapping between model outputs and allowable actions, together with explicit decision boundaries on when and how those outputs may be used in decision-making \cite{Ranisch2025SocialExperiment,Wang2025ChinaGovernance,Ji2025EquityGuard}.
%The third priority is to establish bounded-use governance for action-oriented models. Technically, this requires specifying the mapping between model outputs and allowable actions, together with explicit constraints on when and how those outputs may be used in decision-making \cite{Ranisch2025SocialExperiment,Wang2025ChinaGovernance,Ji2025EquityGuard}.

A governance framework should define: (i) the intended action space supported by the model, such as scenario analysis, prioritization, or resource allocation; (ii) the conditions under which model outputs are considered reliable enough to inform action, including acceptable uncertainty and known failure modes; and (iii) subgroup-specific performance and risk profiles under deployment. These elements should be specified before deployment rather than inferred after harm occurs.

Governance must also be tied to workflow and accountability structure. This includes explicit deployment boundaries, requirements for human oversight at decision points, and auditing mechanisms that track how intervention decisions affect subgroups and population outcomes over time. The goal is not unrestricted adoption, but bounded and accountable use under explicit constraints.

\subsection*{Summary}

Digital medicine is moving from prediction to intervention, where decisions reshape both outcomes and the conditions under which future decisions are made. The three perspectives developed in this article identify what this shift requires: latent health state must be inferred under intervention-dependent observation, population intelligence must reason across coupled scales, and model trustworthiness must be established under action, feedback, and distribution shift.

A PHWM provides a unifying framework for this transition. The required advances are not optional directions but minimal capabilities: representation under action, multi-scale intervention intelligence, and validation with explicit decision boundaries. Together, these capabilities define the technical foundation for moving from prediction to action in digital medicine.

\bibliographystyle{unsrt}
\bibliography{phwm_refs_corrected}

@article{Moor2023GeneralistMedicalAI,
  author  = {Michael Moor and Oishi Banerjee and Zahra Shakeri Hossein Abad and Harlan M. Krumholz and Jure Leskovec and Eric J. Topol and Pranav Rajpurkar},
  title   = {Foundation models for generalist medical artificial intelligence},
  journal = {Nature},
  volume  = {616},
  pages   = {259--265},
  year    = {2023},
  doi     = {10.1038/s41586-023-05881-4},
  url     = {https://www.nature.com/articles/s41586-023-05881-4}
}

@article{
doi:10.1126/scitranslmed.abn6036,
author = {Shuai Xu  and Joohee Kim  and Jessica R. Walter  and Roozbeh Ghaffari  and John A. Rogers },
title = {Translational gaps and opportunities for medical wearables in digital health},
journal = {Science Translational Medicine},
volume = {14},
number = {666},
pages = {eabn6036},
year = {2022},
doi = {10.1126/scitranslmed.abn6036},
URL = {https://www.science.org/doi/abs/10.1126/scitranslmed.abn6036}
}

@article{Richardson2022DigitalHealthEquity,
  author  = {Safiya Richardson and Katharine Lawrence and  Antoinette M. Schoenthaler and Devin Mann},
  title   = {A framework for digital health equity},
  journal = {npj Digital Medicine},
  volume  = {5},
  pages   = {119},
  year    = {2022},
  doi     = {10.1038/s41746-022-00663-0},
  url     = {https://www.nature.com/articles/s41746-022-00663-0}
}

@article{Wilson2024AdvanceDigitalHealthEquity,
  author  = {Sarah Wilson and Clare Tolley and others},
  title   = {Recommendations to advance digital health equity: a systematic review of qualitative studies},
  journal = {npj Digital Medicine},
  volume  = {7},
  pages   = {173},
  year    = {2024},
  doi     = {10.1038/s41746-024-01177-7},
  url     = {https://www.nature.com/articles/s41746-024-01177-7}
}

@article{Ranisch2025SocialExperiment,
  author  = {Robert Ranisch and Joschka Haltaufderheide},
  title   = {Foundation models in medicine are a social experiment: time for an ethical framework},
  journal = {npj Digital Medicine},
  volume  = {8},
  pages   = {525},
  year    = {2025},
  doi     = {10.1038/s41746-025-01924-4},
  url     = {https://www.nature.com/articles/s41746-025-01924-4}
}

@article{Mahajan2025WearableAI,
  author  = {Arjun Mahajan and Kimia Heydari and Dylan Powell},
  title   = {Wearable AI to enhance patient safety and clinical decision-making},
  journal = {npj Digital Medicine},
  volume  = {8},
  pages   = {176},
  year    = {2025},
  doi     = {10.1038/s41746-025-01554-w},
  url     = {https://www.nature.com/articles/s41746-025-01554-w}
}

@article{Ghomrawi2023WearablePostOp,
  author  = {Hassan M. K. Ghomrawi and Megan K. O'Brien and Michela Carter and others},
  title   = {Applying machine learning to consumer wearable data for the early detection of complications after pediatric appendectomy},
  journal = {npj Digital Medicine},
  volume  = {6},
  pages   = {148},
  year    = {2023},
  doi     = {10.1038/s41746-023-00890-z},
  url     = {https://www.nature.com/articles/s41746-023-00890-z}
}

@article{Gupta2023TrajectoryPatterns,
  author  = {Varsha Gupta and Sokratis Kariotis and Allan Lawrie and others},
  title   = {Unsupervised machine learning to investigate trajectory patterns of COVID-19 symptoms and physical activity measured via the MyHeart Counts app and smart devices},
  journal = {npj Digital Medicine},
  volume  = {6},
  pages   = {239},
  year    = {2023},
  doi     = {10.1038/s41746-023-00974-w},
  url     = {https://www.nature.com/articles/s41746-023-00974-w}
}

@article{Bladon2025PassiveData,
  author  = {Si{\^a}n Bladon and Emily Eisner and Sandra Bucci and others},
  title   = {A systematic review of passive data for remote monitoring in psychosis and schizophrenia},
  journal = {npj Digital Medicine},
  volume  = {8},
  pages   = {62},
  year    = {2025},
  doi     = {10.1038/s41746-025-01451-2},
  url     = {https://www.nature.com/articles/s41746-025-01451-2}
}

@article{Kalmady2024PopulationECG,
  author  = {Sunil Vasu Kalmady and Amir Salimi and Weijie Sun and others},
  title   = {Development and validation of machine learning algorithms based on electrocardiograms for cardiovascular diagnoses at the population level},
  journal = {npj Digital Medicine},
  volume  = {7},
  pages   = {133},
  year    = {2024},
  doi     = {10.1038/s41746-024-01130-8},
  url     = {https://www.nature.com/articles/s41746-024-01130-8}
}

@article{Low2020CancerSensors,
  author  = {Carissa A. Low},
  title   = {Harnessing consumer smartphone and wearable sensors for clinical cancer research},
  journal = {npj Digital Medicine},
  volume  = {3},
  pages   = {140},
  year    = {2020},
  doi     = {10.1038/s41746-020-00351-x},
  url     = {https://www.nature.com/articles/s41746-020-00351-x}
}

@article{Lampos2021OnlineSearch,
  author  = {Vasileios Lampos and Maimuna S. Majumder and Elad Yom-Tov and others},
  title   = {Tracking COVID-19 using online search},
  journal = {npj Digital Medicine},
  volume  = {4},
  pages   = {17},
  year    = {2021},
  doi     = {10.1038/s41746-021-00384-w},
  url     = {https://www.nature.com/articles/s41746-021-00384-w}
}

@misc{LeCun2022PathAMI,
  author       = {Yann LeCun},
  title        = {A Path Towards Autonomous Machine Intelligence},
  year         = {2022},
  note         = {Version 0.9.2, 2022-06-27; OpenReview position paper},
  url          = {https://openreview.net/pdf?id=BZ5a1r-kVsf}
}

@article{Ha2018WorldModels,
  author  = {David Ha and J{\"u}rgen Schmidhuber},
  title   = {World Models},
  journal = {arXiv preprint arXiv:1803.10122},
  year    = {2018},
  url     = {https://arxiv.org/abs/1803.10122}
}

@inproceedings{DBLP:conf/nips/HaS18,
  author       = {David Ha and
                  J{\"{u}}rgen Schmidhuber},
  editor       = {Samy Bengio and
                  Hanna M. Wallach and
                  Hugo Larochelle and
                 others},
  title        = {Recurrent World Models Facilitate Policy Evolution},
  booktitle    = {Advances in Neural Information Processing Systems},
  pages        = {2455--2467},
  year         = {2018},
}

@article{Kaelbling1998POMDP,
  author  = {Leslie Pack Kaelbling and Michael L. Littman and Anthony R. Cassandra},
  title   = {Planning and acting in partially observable stochastic domains},
  journal = {Artificial Intelligence},
  volume  = {101},
  number  = {1-2},
  pages   = {99--134},
  year    = {1998},
  doi     = {10.1016/S0004-3702(98)00023-X},
  url     = {https://people.csail.mit.edu/lpk/papers/aij98-pomdp.pdf}
}

@inproceedings{Mao2024GraphFoundationModels,
  author    = {Haitao Mao and Zhikai Chen and Wenzhuo Tang and others},
  title     = {Position: Graph Foundation Models Are Already Here},
  booktitle = {Proceedings of the International Conference on Machine Learning},
  pages     = {34670--34692},
  year      = {2024},
  url       = {https://proceedings.mlr.press/v235/mao24a.html}
}

@inproceedings{Valmeekam2023PlanBench,
  author    = {Karthik Valmeekam and
                  Matthew Marquez and
                  Alberto Olmo Hernandez and
                  others},
  title     = {PlanBench: An Extensible Benchmark for Evaluating Large Language Models on Planning and Reasoning about Change},
  booktitle = {Advances in Neural Information Processing Systems},
  pages = {38975 - 3898},
  year      = {2023},
  url       = {https://proceedings.neurips.cc/paper_files/paper/2023/hash/7a92bcdede88c7afd108072faf5485c8-Abstract-Datasets_and_Benchmarks.html}
}

@inproceedings{Qiao2024WorldKnowledgeModel,
  author    = {Shuofei Qiao and Runnan Fang and Ningyu Zhang and others},
  title     = {Agent Planning with World Knowledge Model},
  booktitle = {Advances in Neural Information Processing Systems},
  articleno = {3646},
  pages = {114843 - 114871},
  year      = {2024},
  url       = {https://proceedings.neurips.cc/paper_files/paper/2024/hash/d032263772946dd5026e7f3cd22bce5b-Abstract-Conference.html}
}

@article{Wang2025ChinaGovernance,
  author  = {Minmin Wang and Xinran Lu and Yeting Du and others},
  title   = {Digital health governance in China by a whole-of-society approach},
  journal = {npj Digital Medicine},
  volume  = {8},
  pages   = {496},
  year    = {2025},
  doi     = {10.1038/s41746-025-01876-9},
  url     = {https://www.nature.com/articles/s41746-025-01876-9}
}

@article{Ji2025EquityGuard,
  author  = {Yuelyu Ji and Wenhe Ma and Sonish Sivarajkumar and others},
  title   = {Mitigating the risk of health inequity exacerbated by large language models},
  journal = {npj Digital Medicine},
  volume  = {8},
  pages   = {246},
  year    = {2025},
  doi     = {10.1038/s41746-025-01576-4},
  url     = {https://www.nature.com/articles/s41746-025-01576-4}
}

\end{document}